%% file: wbir2014.tex
\documentclass{llncs}
\pdfoutput=1 
\usepackage{llncsdoc}
\usepackage{times}
\usepackage{graphicx}
\usepackage{amsmath}
\usepackage{amssymb}
\input{demPreamble}

\usepackage{cleveref}
\Crefname{equation}{Eq.}{Eqs.}
\Crefname{figure}{Fig.}{Figs.}
\Crefname{proposition}{Prop.}{Props.}
\usepackage{mathtools}

\begin{document}

\title{Probabilistic Diffeomorphic Registration: \\ Representing Uncertainty\thanks{This work was supported by grants NIH P41EB015898, R01CA138419 and NSF EECS-1148870.}\vspace{-10pt}}
\author{Demian Wassermann\inst{1,3} \and Matt Toews\inst{1} \and Marc Niethammer\inst{4} \and William Wells III\inst{1,2}
}
\institute{SPL, Brigham and Women’s Hospital, Harvard Medical School, Boston, MA, USA
\and CSAIL, Massachusetts Institute of Technology, Boston, MA, USA
\and EPI Athena, INRIA Sophia Antipolis-M\'editerran\'ee, Sophia Antipolis, France
\and School of Medicine, University of North Carolina, Chapel Hill, NC, USA}
\maketitle

\begin{abstract}
\vspace{-20pt}
This paper presents a novel mathematical framework for representing uncertainty in large deformation diffeomorphic image registration.
The Bayesian posterior distribution over the deformations aligning a moving and a fixed image is approximated via a variational formulation. A stochastic differential equation (SDE) modeling the deformations as the evolution of a time-varying velocity field leads to a prior density over deformations in the form of a Gaussian process. This permits estimating the full posterior distribution in order to represent uncertainty, in contrast to methods in which the posterior is approximated via Monte Carlo sampling or maximized in maximum a-posteriori (MAP) estimation. The framework is demonstrated in the case of landmark-based image registration, including simulated data and annotated pre and intra-operative 3D images.\vspace{-15pt}
\end{abstract}
\section{Introduction}\vspace{-10pt}
\input{introduction}

\section{Methods}\vspace{-10pt}
\input{methods}

\section{Experiments}\vspace{-10pt}
\input{experiments}

\section{Discussion and Conclusion}\vspace{-10pt}
\input{discussion}

\vspace{-10pt}
{
\bibliographystyle{splncs}
}
\bibliography{dw_biblio_abb,mt_biblio}
\end{document}

%% file: demPreamble.tex
\usepackage{etex}
\usepackage{ifthen}
\usepackage{xargs}

\usepackage{xcolor}

\usepackage{amssymb}
\usepackage{amsmath}

\usepackage{graphicx}
\usepackage{multirow}
\usepackage{enumerate}
\usepackage[center,scriptsize]{subfigure}

\usepackage[font=small,labelfont=bf, skip=-10pt,
               labelsep=endash]{caption}

\usepackage{array}
\newcolumntype{V}{ c }

\ifthenelse{\isundefined{\citep}}{\newcommand{\citep}[2][]{\cite{#2}}}{}
\ifthenelse{\isundefined{\citet}}{\newcommand{\citet}[2][]{\cite{#2}}}{}

\def\url#1{}

\newcommand{\eg}{e.g.\ }

\newcommand{\CHANGED}[1]{{\color{black}#1}}
\renewcommand{\Re}{\mathbb{R}}
\newcommand{\rv}[1]{\ensuremath{\boldsymbol{\mathbf{#1}}}}
\newcommand{\E}[2][ ]{\ensuremath{\left\langle#2\right\rangle_{#1}}}
\newcommand{\KL}[1]{\ensuremath{\textrm{KL}\left[#1\right]}}

\newcommand{\Covar}[2][ ]{\ensuremath{\operatorname{cov}_{#1}\left[#2\right]}}

\def\GP{\ensuremath{\mathcal{GP}}}
\def\G{\ensuremath{\mathcal{G}}}

\def\G{\ensuremath{\mathcal{G}}}

\def\transp{{\mathstrut\intercal}}

\newcommand{\partiald}[2][ ]{\ensuremath{\partial_{#1}{#2}\!}}
\newcommand{\jac}[2][ ]{\ensuremath{\mathsf D_{#2}^{#1}\!}}

\DeclareMathOperator{\trace}{tr}

\DeclareMathOperator*{\argmin}{argmin}

\DeclareMathOperator{\id}{id}
\DeclareMathOperator{\SPD}{SPD}
\DeclareMathOperator{\Id}{\operatorname{Id}}

\DeclareMathOperator{\V}{vec}
\DeclareMathOperator{\deformed}{\circ}

%% file: introduction.tex
Deformable image registration seeks to identify a deformation field that aligns two images, and is a key component of image analysis applications such as computational anatomy~\cite{Dupuis:1998ur,Joshi-Miller:2000}. An important body of literature focuses on deformations in the form of diffeomorphisms~\citep{Ashburner:2011cp,Beg-Miller-etal:2005,Joshi-Miller:2000}, one-to-one mappings between image coordinate systems that are smooth and invertible. These properties help in ensuring biologically plausible deformations, and avoiding phenomena such as folding or tearing that may occur in non-diffeomorphic registration approaches~\citep{Simpson:2013ia}.

While a good deal of literature has focused on identifying optimal diffeomorphic registration solutions, it would be useful to quantify the inherent uncertainty in these solutions when interpreting the results of registration. Quantification of deformable registration uncertainty, particularly at point locations throughout the image, remains an open problem. The Bayesian approach quantifies probabilistic uncertainty via a posterior distribution over deformations conditioned on image data. Estimating the full posterior in the case of large deformation diffeomorphisms is desirable but computationally challenging, and has typically been avoided. Simpson et al. propose a Bayesian variational framework based on small deformation kinematics~\cite{Simpson:2013ia}, however this does not address the general case of large deformations. Markussen proposed a stochastic differential equation (SDE) model for large deformations, however only the computation of the maximum a posteriori deformation is provided lacking the estimation of a distribution on the deformations~\cite{Markussen:2007hv}Alternatively, the posterior may be investigated via sampling methods, e.g. Markov chain Monte Carlo (MCMC)~\cite{risholm2013bayesian} or Hamiltonian Monte Carlo~\cite{zhang2013bayesian}.

This paper introduces a novel mathematical framework that allows representing and computing of the full Bayesian posterior in the case of large deformation diffeomorphisms. Our framework considers a SDE modeling the deformation field as the evolution of a time-varying velocity field, with additive noise in the form of a Wiener process. A Gaussian process (GP) density results from a locally linear approximation of the SDE and taking the initial deformation field to be Gaussian process distributed. Deformation field uncertainty is quantified by the point-wise covariance of the deformation field throughout the image, and can be summarized, e.g., via the Frobenius norm of the covariance (FC). This can be pictured through the following example: if the FC at a point approaches $0$, the marginal density of the transform approaches an impulse function denoting the existence of a single probable solution. On the other hand, when FC is large, the density becomes ``broader'' denoting a larger set of solutions with high probability at that point. Hence the point-wise FC is a model of uncertainty. Experiments demonstrate our framework in the context of landmark correspondences, where a heteroscedastic model accounts for variable uncertainty in landmark localization. This is particularly useful when estimates of landmark localization variability are available.

%% file: methods.tex
\subsection{Variational Approximation to Registration}\label{sec:varap-reg}\vspace{-8pt}
We start by posing the registration problem in a probabilistic framework. Let $M$ and $F$ be moving and fixed objects with domains in $\Omega_M$ and $\Omega_F$ respectively, and let $\phi:\Omega_M\mapsto \Omega_F$ be a mapping between the two. The registration problem seeks a posterior probability density over mappings $\phi$ conditioned on data $(M,F)$, which is expressed via Bayes theorem as
\begin{equation}\label{eq:bayes-reg}
  p(\rv \phi | M, F)  = p(\rv \phi)p(M, F|\rv \phi) / p(M,F).
\end{equation}
In~\Cref{eq:bayes-reg}, $p(\rv \phi)$ is a prior density over $\rv \phi$ embodying geometrical constraints such as smoothness. $p(M, F|\rv \phi)$ is the data attachment factor or \CHANGED{likelihood of the map $\phi$ relating $F$ and $M$}. E.g. the probability that $M$ deformed by $\phi$, which we note $\phi\deformed M$, is similar to $F$. Finally, $p(M,F)$ is a normalizing constant.

The direct calculation of the posterior density  $p(\rv \phi |M, F)$ is a difficult problem. Hence, we use a variational method to estimate a distribution $q(\rv \phi)$ (abbreviated as $q$) that is close to $p(\rv \phi |M, F)$ in the sense of the Kullback-Leibler divergence ($\KL{\cdot\|\cdot}$). Specifically, we seek $q$ minimizing:
\begin{equation}\label{eq:KL-approach}
\begin{aligned}
  \KL{q\|p(\rv \phi | M, F)} &= 
\KL{ q\|p(\rv \phi) } - \int \log p(M,F|\rv \phi) dq(\rv \phi) + \log p(M,F).
\end{aligned}
\end{equation}

In the registration literature the data attachment factor $p(M, F|\rv \phi)$ is typically modeled using a measure of similarity between the registered objects: $m:\Omega_F\times\Omega_F\mapsto\Re$ which is minimal when two objects are exactly the same and grows as they become different. Adopting the equality $-\log p(M, F|\rv \phi) = m(\phi\deformed M, F)$,
\Cref{eq:KL-approach} may be rewritten as:
\begin{multline}\label{eq:KL-approach-entropy}
  \KL{q\|p(\rv \phi | M, F)} = \KL{ q\|p(\rv \phi) } 
  + \E[q]{m(\rv \phi \deformed M, F)} + \log {p(M,F)},
\end{multline}
where $\E[q]{ m(\rv \phi \deformed M,F)}$ is the expected value of $m$ with respect to the density $q$. There are two main differences of this formulation with respect to common diffeomorphic registration approaches~\citep[\eg]{Ashburner:2011cp,Beg-Miller-etal:2005}. First, instead of seeking a single optimal deformation $\phi$, e.g. the maximum a-posteriori (MAP) solution in the Bayesian formulation, we seek to obtain the full distribution $q(\rv \phi)$. In this way, we obtain both the MAP deformation $\phi$ in addition to the uncertainty at any given point in space, which can be calculated from $q(\rv \phi)$. Second, we obtain $q(\rv \phi)$ by minimizing the data attachment term over a weighted combination of \emph{all possible deformation fields in the family of $\phi$} instead of only at a single deformation $\phi$.

\subsection{Probabilistic Diffeomorphic Deformations}\label{sec:prob-diff-def}\vspace{-8pt}

The variational approximation to $p(\rv \phi|M,F)$ described in the previous section requires a parameterization for $q(\rv \phi)$ over which~\Cref{eq:KL-approach-entropy} can be minimized. In this section we derive a novel parameterization in the form of a Gaussian Process (GP). The theoretical basis for our derivation lies in a stochastic interpretation of the work of~\citet{Dupuis:1998ur}, common to many diffeomorphic registration approaches~\citep[see \eg]{Joshi-Miller:2000}. Here, we begin by outlining the relevant elements of this work, then we present our derivation in three propositions and their proofs, with our primary contributions being in Propositions 2 and 3.

Following the work of \citet{Dupuis:1998ur}, many diffeomorphic deformation formulations seek an optimal registration solution $\phi$, \eg the MAP deformation in \Cref{eq:bayes-reg}, by constraining the map $\phi$ to be the solution at $t=1$ of the ordinary differential equation (ODE)
\begin{equation}\label{eq:long-diff} 
  \tfrac d {dt} \phi_t( x) =  v_t(\phi_t( x)), \quad \phi_0( x)= x,\quad t\in[0, 1].
\end{equation}
and setting $\phi_1$ in~\Cref{eq:long-diff} to minimize
\newcommand{\Ev}{\ensuremath{E_{ v}( v)}}\newcommand{\Ep}{\ensuremath{E_\phi(\phi; M, F)}}
\begin{equation}\label{eq:minimal-v}
  E(\phi; M, F) = \Ev + \Ep, \quad \Ev = \frac 1 2 \int_0^1\int_{\Omega_m} \|L  v_t( x)\|^2_2 d x dt.
\end{equation}
From the two terms of $E$, $\Ev$ regularizes the evolution of the time-varying velocity field and $\Ep$ drives $E$ such that the deformed moving object, $\phi_1\deformed M$, becomes as similar to $F$ as possible. The regularization term $\Ev$, is driven by $L$, a linear differential operator. The key insight here is that given suitable $L$ \CHANGED{according to \citep{Dupuis:1998ur}}, \Cref{eq:minimal-v} restricts $\phi_t( x)$ defined as in \Cref{eq:long-diff} to the space of diffeomorphisms~\cite{Dupuis:1998ur}. From this, we make the following propositions:

\begin{proposition}\label{prop:v-gp}
  Under a probabilistic interpretation, the regularization term $\Ev$ in \Cref{eq:minimal-v} corresponds to the negative logarithm of the GP prior on the stochastic velocity field $\rv v$
  \begin{equation}\label{eq:v-gp}
    p( \rv v)= \GP( 0, \Sigma_{{ts}}( x,  y)),\quad \Sigma_{{ts}}( x,  y)\in \SPD^d
  \end{equation}
  where the covariance function $\Sigma_{{ts}}( x,  y)$, representing the relation between the point $ x$ at time $t$ and the point $ y$ at time $s$, is determined by the operator $L$ in \Cref{eq:minimal-v}.
\end{proposition}

%

\begin{proposition}\label{prop:sde} Interpreting the energy $\Ev$ in \Cref{eq:minimal-v} as the negative logarithm of the density of a stochastic process $\rv v$ induces a random process $\rv \phi$ with density $p(\rv \phi)$ on the deformation field of \Cref{eq:long-diff} that is a solution of the stochastic differential equation (SDE)
  \begin{align}
    d \rv \phi_t( x) &=  v_t(\rv \phi_t( x))dt + \sqrt{\Sigma_{t}(\rv \phi_t( x))}d\rv W_t\label{eq:prop-sde}, \\
      \rv W_t &\sim \GP( 0, \Theta_{{ts}}( x,  y)),\quad \Theta_{{ts}}( x,  y)=\min(t,s)\Id\label{eq:prop-sde-wiener},
  \end{align} 
  where $\sqrt{\Sigma}$ is the square root matrix $\sqrt{\Sigma}\sqrt{\Sigma}^\transp  = \Sigma$; the GP $\rv W_t\in \Re^d$ is called Brownian motion or a Wiener process~\cite{Kloeden:1992vt}; $ v_t( x) \in \Re^d$ is a deterministic velocity field like in \Cref{eq:long-diff}; and  $\Sigma_{t}( x)\triangleq\Sigma_{tt}( x,  x)\in \SPD^d$~\footnote{$SPD^d$: set of symmetric positive definite matrices of dimension $d$}, the covariance of the probabilistic prior defined in \Cref{prop:v-gp}, is a consequence of \Cref{eq:minimal-v}.
\end{proposition}


\begin{proposition}\label{prop:sde-gp}
  For the stochastic process $\phi$ with density $p(\phi)$, defined in \cref{prop:sde}, the mean $\bar \phi$ and covariance $\Lambda$ functions are solutions of the deterministic ODEs
  \begin{subequations}\label{eq:phi-de-param}
  \begin{align}
    \tfrac{d}{dt}\bar \phi_t( x) &=\E[p]{{ v}_t\left(\vec \phi_t( x)\right)}\label{eq:phi-de-param-mean}
\\
\Lambda_{{ts}}( x,  y) &= \Covar[p]{\rv \phi_t( x), \rv \phi_s( y)}= \E[p]{\rv \phi_t( x)\rv \phi_s^\transp ( y)} - \bar \phi_t( x)\bar\phi_s^\transp ( y)\\
    \tfrac{d}{dt}\E[p]{\rv \phi_t( x)\rv \phi_s^\transp ( y)}&=\E[p]{{ v}_t(\rv \phi_t( x))\rv \phi_s^\transp( y)} + \E[p]{\rv \phi_t( x) v_s^\transp(\rv \phi_s( y))}\label{eq:phi-de-param-2ndm}\\
    &+\E[p]{{\Sigma}_{ts}(\rv\phi_t( x), \rv\phi_s( y))}. \notag
\end{align}
  \end{subequations}
\CHANGED{Moreover,  up to a first order approximation:
\begin{equation}\label{eq:prop-gp-approx}
      p(\rv \phi)  = \GP(\bar \phi_t( x), \Lambda_{ts}( x,  y)),
    \end{equation}}
\end{proposition}

\vspace{-25pt}The proofs for Propositions 1-3 are as follows:

\textbf{Proof of \Cref{prop:v-gp}}  This proposition has been proven by Joshi et. al.\cite{Joshi-Miller:2000}, here we provide a sketch of the relevant points. We start by relating $\Ev$ in \Cref{eq:minimal-v} to a probability density on velocity fields as stochastic processes $\rv v$, $p( \rv v)$:
  \begin{equation}\label{eq:v-prior-cont}
    -\log(p(\rv v)) =  \Ev + const = \frac 1 2 \int_0^1\int_{\Omega_M} \|L \rv v_t( x)\|^2_2 d x dt + const,
  \end{equation}
  To show that $p(\rv v)$ is a stochastic process with a particular distribution, we need to prove that any finite sample of the domain $\Omega_M \in \Re^d$ has the same parametric distribution~\citep{Rasmussen-Williams:2006}. We take $N$ samples $X \in \Re^{N\times d}$ in space and $ t \in [0, 1]^N$ in time, and let  $\rv V_{ij}= \left[ \rv v_{ t_i}\left(X_i\right)\right]_j \in \Re^{N\times d}$. Then we rewrite \Cref{eq:v-prior-cont} as
  \begin{equation}\label{eq:log-p-gauss}
    -\log(p(\rv V)) = \tfrac 1 2 \left( L\V \rv V\right)^\transp  L\V \rv V + \text{const}
    = \tfrac 1 2  \V \rv V^\transp   L^\transp   L\V \rv V + \text{const},
  \end{equation}
  where $ L$ is the matrix such that $[ L\rv V_{i\cdot}]=[L \rv v_{t_i}( X_i)]_j$. \Cref{eq:log-p-gauss} is recognisable as the log probability of a centered multivariate Gaussian with covariance $C= (L^\transp L)^{-1}$ and, therefore $ \rv v( x)$ is a GP. The covariance function $\Sigma_{{ts}}( x, y)$ can be calculated as the matrix Green's function of the operator $L$~\cite{Joshi-Miller:2000}; specifically, if $ x,  y \in \Re^d$, $\Sigma_{{ts}}( x,  y) \in \SPD^{d}$  where $[\Sigma_{{ts}}( x, y)]_{ij}$ is the covariance between $ x_i$ at time $t$ and $ y_j$ at time  $s$.

  This shows that for a given velocity field ${\rv v}( x)$, random perturbations according to the regularization term $\Ev$ in \Cref{eq:minimal-v} or prior \Cref{eq:v-prior-cont} follow a GP, therefore the velocity fields according to $\Ev$ in~\Cref{eq:minimal-v} have the density specified in~\Cref{eq:v-gp} proving \Cref{prop:v-gp}.

  \textbf{Proof of \Cref{prop:sde}} A formal proof of \Cref{prop:sde} is beyond the scope of this paper. Instead, using \Cref{prop:v-gp} we argue its validity and provide appropriate references. In \Cref{prop:v-gp} we characterized the density of random perturbations of velocity fields according to \Cref{eq:minimal-v}. Adding such random perturbations to \Cref{eq:long-diff} leads to \Cref{eq:prop-sde}.

  The second term in \Cref{eq:prop-sde} comes from considering the velocity fields $v_t$ of \Cref{eq:long-diff} as a stochastic process according to \Cref{prop:v-gp}. We achieve this by perturbing the right hand side of \Cref{eq:long-diff} with noise. The factor $\rv W_t\in \Re^d$ is white noise, which multiplied by $\sqrt \Sigma_{t}$ is a centered Gaussian random variable with covariance $\Sigma_{t}$, a sample drawn from~\Cref{eq:v-gp}. We noted the \emph{stochastic} velocity field in \Cref{eq:prop-sde} ${\rv v}$ to distinguish it from the deterministic one $v$. \Cref{eq:prop-sde} ceases to be an ODE as a sample path of $\rv W_t$ is almost surely not differentiable. Alternatively, using the It\=o interpretation of \Cref{eq:prop-sde} leads to the SDE in~\Cref{eq:prop-sde}, whose solution is the density on $\rv \phi$~\cite[Chap. 8]{Kloeden:1992vt}.

  \textbf{Proof of \Cref{prop:sde-gp}} The ODE for the mean of the stochastic process $\rv \phi$, \Cref{eq:phi-de-param-mean}, is obtained by calculating the expectation on both sides of \Cref{eq:prop-sde}. It is a consequence of the linearity of the expected value and the derivative operator and the definition of $\rv W_t$ as a zero-centered Wiener process in \Cref{eq:prop-sde-wiener}.

To obtain the ODE for the second moment of $\rv \phi$, shown in~\Cref{eq:phi-de-param-2ndm}, we use the It\=o product rule \citep{Kloeden:1992vt} to obtain an expression for $d(\rv \phi_t( x)\rv \phi_s^\transp( y))$ and  substitute it in \Cref{eq:prop-sde} obtaining
\begin{equation}
  \begin{aligned}
  &\E[p]{d(\rv \phi_t( x)\rv \phi_s^\transp( y))}=d \E[p]{\rv \phi_t( x)\rv \phi_s^\transp( y)}
  =\E[p]{v_t(\rv \phi_t ( x))dt\rv \phi_s^\transp( y)} + \E[p]{\sqrt{\Sigma}_t(\rv \phi_t ( x))d\rv W_t\rv \phi_s^\transp( y)}\\
  &\quad+\E[p]{\rv \phi_t( x)(v_s(\rv \phi_s ( y)))^\transp ds} + \E[p]{\rv \phi_t( x)(\sqrt{\Sigma}_t(\rv \phi_s ( y))d\rv W_s)^\transp}\\
  &\quad+\E[p]{\bigg(v_t(\rv \phi_t( y))dt+\sqrt{\Sigma}_t(\rv \phi_t( y))d\rv W_t\bigg)\bigg(v_s(\rv \phi_s( y))ds+\sqrt{\Sigma}_s(\rv \phi_s( y))d\rv W_s\bigg)^\transp},
  \end{aligned}
\end{equation}
which, using the It\=o identities for expected values of differentials~\citep{Kloeden:1992vt} results in \Cref{eq:phi-de-param-2ndm}.

Obtaining a parametric form of the density of $\phi$, $p(\phi)$, satisfying the SDE~(\ref{eq:prop-sde}) in the general case is an open problem and a wide field of study. However, in the case where the drift $v$ and diffusion coefficient $\sqrt{\Sigma}$ are linear functions on their time and location parameters, and the initial condition $\phi_{t=0}$ is a GP, $\phi_t( x)$ is known to be a GP~\cite{Kloeden:1992vt}. With this purpose we define a locally linearized (LL) $v$ and $\sqrt{\Sigma}$ centered at $t_0,  x_0$~\cite{Biscay:1996ho}:
\begin{subequations}
\begin{align}
v_t( x) &\approx v_{t_0}( x_0) + \partiald[t]{v_{t_0}}( x_0)(t-t_0)
+ \jac[ x]{v_{t_0}} ( x_0)( x -  x_0) \label{eq:approx-vs} \\
{\sqrt{\Sigma}}_{t}( x) &\approx  {\sqrt{\Sigma}}_{t_0}( x_0)+ \partiald[t]{{\sqrt{\Sigma}}_{t_0}}( x_0)(t-t_0) 
+ \sum_i \partiald[ x_i]{{\sqrt{\Sigma}}_{t_0}} ( x_0)( x -  x_0)_i \label{eq:approx-ss}
\end{align}
\end{subequations}
where $\jac[ x]{v_t}$ is the Jacobian of $v_t( x)$ w.r.t. $ x$ and $\partiald[t]{v_t}$ its partial derivative w.r.t. $t$. Considering that $L$ is assumed time-invariant in \Cref{prop:v-gp}, the time derivative of $\sqrt{\Sigma}$  in \Cref{eq:approx-ss} is equal to $ 0$. Then, using the LL equations \Cref{eq:approx-vs,eq:approx-ss}, we approximate \Cref{eq:prop-sde} as
\begin{gather}\label{eq:phi-ll}
  d\rv \phi_t( x) \approx (A_t \rv \phi_t( x) +  a_t)dt + \left(\sum_i S_t^i\rv\phi_t( x)_i+R_t\right)d\rv W_t\\
  \begin{alignedat}{2}\label{eq:phi-ll-cts}
  A_t &\triangleq \jac[ x]{v_{t_0}} ( x_0) &\quad
   a_t &\triangleq -\jac[ x]{v_{t_0}}( x_0) x_0 + \partiald[t]{v_{t_0}}( x_0)(t-t_0)+ v_{t_0}( x_0)\\
  S^i_t &\triangleq\partiald[ x_i]{{\sqrt{\Sigma}}_{t_0}} ( x_0)&\quad
  R_t &\triangleq-\sum_i S^i_t\cdot( x_0)_i + {\sqrt{\Sigma}}_{t_0}( x_0).
\end{alignedat}
\end{gather}
The LL approximations in \Cref{eq:phi-ll,eq:approx-vs} lead to an approximation of the mean function of $\rv \phi$, $\bar \phi$, by the solution of the ODE
\begin{equation}\label{eq:phi-ll-mean}
  \tfrac{d\bar \phi_t( x)}{dt} \approx v_t(\bar \phi_t( x)) \text{ where } v_t( x) \approx (A_t\bar \phi_t( x) +  a_t),
\end{equation}
and its second moment $\E{\rv \phi_t( x)\rv \phi_s( y)^\transp}$ when $t=s$ by
  \begin{equation}\label{eq:phi-ll-2ndm}
    \begin{split}
      \tfrac{d\E{\rv\phi_t( x) \rv\phi_t^\transp ( y)}}{dt}
      \approx A_t\E{\rv\phi_t( x)\rv\phi_t^\transp( y) }+ \E{\rv\phi_t( x)\rv\phi_t^\transp( y) }A_t'^\transp 
   + { a}_t\bar\phi_t^\transp( y) + \bar\phi_t( x){ a}_t'^\transp\\ +\sum_{ij} S_t^i\E{\rv\phi_t( x)\rv\phi_t^\transp( y) }\left(S_t'^j\right)^\transp 
  +\left(\sum_i S_t^i\bar\phi_t( x)_i\right)R_t'^\transp +R_t\left(\sum_i S_t'^i\bar\phi_t( y)_i\right)^\transp+R_tR_t'^\transp,
  \end{split}
\end{equation}
where $A'_t$; $ a'_t$; $S'^i_t$; and $R'_t$ are the same as $A_t$; $ a_t$; $S^i_t$; and $R_t$ in \Cref{eq:phi-ll-cts} substituting $ y$ and $ y_0$ for $ x$ and $ x_0$.

As long as the initial condition $\rv \phi_{t=0}$ is a GP, the linear approximation of $\rv \phi_t$ is a GP uniquely determined by $\bar \phi$ and $\Lambda$~\citep{Kloeden:1992vt}. Then, given a set of stochastic velocity fields  $ v_{0}\ldots  v_{t_{M-1}}$ with $t_0=0$ and $t_{M-1}=1$, the parameters of the stochastic process representing the transform $\rv \phi$ are obtained integrating~\Cref{eq:phi-ll-mean,eq:phi-ll-2ndm} with the initial conditions $\rv \phi_{t=0}\sim\GP(\bar \phi_{t=0}, \Lambda_{t=0})$. Having characterized stochastic transformations representing a diffeomorphic deformation, we are in position to formulate our probabilistic diffeomorphic registration algorithm.

\subsection{Probabilistic Diffeomorphic Registration}\label{sec:registration}\vspace{-8pt}
\CHANGED{ The stochastic diffeomorphic deformation model of \Cref{sec:prob-diff-def} leads to a GP approximation on deformation fields,
whose parameters are determined by $v$ and $\Sigma$; we use this model as $q(\rv \phi_1)$, our variational distribution. In this section, we show how to compute the parameters of $q(\rv \phi_1)$ minimizing \Cref{eq:KL-approach} for a particular registration problem.}
Taking the approach of \cite{Ashburner:2011cp}, we focus on operators $L$ for the energy \Cref{eq:minimal-v} regularizing in space but not in time. Due to the time-independent regularization, the prior on velocity fields of $\rv \phi$ derived with \Cref{prop:v-gp} is the joint probability of the fields at each time $t$: $p(\rv \phi_1)=\prod_0^1 p(\rv v_t)^{dt}$ with $p(\rv v_t)\sim \GP( 0, \Sigma_{0})$. Then, we rewrite leftmost term of \Cref{eq:KL-approach} as $\KL{q\|p(\rv \phi_1)}=\int_0^1 \KL{q(\rv v_t)|p(\rv v_t)}dt$. We parameterize each stochastic velocity field  $\rv v_t$ by a $N$-point set represented as a matrix $X_t\in\Re^{N\times d}$ rendering its mean equivalent to a spline model~\citep{Rasmussen-Williams:2006}. This sets the distributions of the discretized velocity field prior to $p(\V  \rv v_t|X_t)=\G( 0, S_{t=0}|X_t)$. \CHANGED{As in usual LDDMM approaches, we keep the $L$ operator, hence the covariance $S$, fixed. Hence, the parameterized form of variational approximation to the posterior of the velocity fields becomes $q(\V \rv v_t |X_t)=\G( \mu_{t}, S_{t=0}|X_t)$}. Due to GP properties given the mean and covariance functions for the GP, we can characterize the mean and covariance for the discretized velocity field as, $\mu_{t}(X)=\V v_t(X)$ and $[S_{t=0}(X)]_{di+k,dj+l}=[\Sigma_{t=0}(X_i,X_j)]_{kl}, i,j=1\ldots N, k,l=0\ldots d-1$~\cite{Rasmussen-Williams:2006}. This leads to an objective which we minimize to obtain $q(\rv \phi_1)$ representing the registration problem:
\begin{subequations}\label{eq:disc-KL-approach}
\begin{gather}
  \mathcal E(q(\rv \phi_1)) = \KL{q(\rv \phi_1)\|p(\rv \phi_1)}+\langle m(\rv \phi_1 \deformed M , F)\rangle_q + \log p(M, F)\\
  \KL{q(\rv \phi_1)\|p(\rv \phi_1)}=\int_0^1  \mu_{t}^\transp S_{t=0}^{-1} \mu_{t} dt, \text{ s.t. } \tfrac{d\bar \phi_t}{dt} \approx  \mu_t(\bar \phi_t),  \bar \phi_0 = \id.
\end{gather}
\end{subequations}
Using the ideas of~\citep{Ashburner:2011cp}, \Cref{eq:disc-KL-approach} can be minimized through geodesic shooting~\citep{Beg-Miller-etal:2005}, i.e. it depends only on $M$, $F$ and $ \mu_{t=0}$. The shooting equations for the proposed probabilistic diffeomorphic registration can be derived using \Cref{eq:disc-KL-approach} in combination with the evolution equation based on the most probable velocity field $ \mu$. In fact, the problem formulation equations, shown in \Cref{eq:long-diff}, stay the same as in~\citep{Ashburner:2011cp}, only the final condition $E_\phi$ changes, which is then “warped” to $t = 0$ for a gradient descent with respect to the initial velocity $ \mu_{t=0}$ leading to the objective function of $q(\rv \phi_1)$ parameterized on $\mu_{t=0}$
\begin{equation}\begin{gathered}\label{eq:disc-KL-approach-momenta}
  \argmin_{\mu_{t=0}} \mathcal E( q_{\mu_{t=0}}(\rv\phi)) = \tfrac 1 2  \mu_{t=0}^\transp S^{-1}_{t=0} \mu_{t=0}^\transp  + \langle m(\rv \phi_1 \deformed M , F)\rangle_q  - \log p(M, F).
\end{gathered}\end{equation}
Up to this point the framework we presented is general for cases where $M$ and $F$ are images or landmarks. Henceforth, we specialize the treatment of the registration problems for the landmark case where $M$ and $F$ are matrices in $\Re^{N\times d}$; $m(M, F) = \|M-F\|^2_2$; and the random variable $\rv \Phi_t = \V (\rv \phi_t \deformed M) \triangleq \V \rv \phi_t(M)$. This allows us to rewrite 
\begin{equation}\label{eq:disc-KL-data}
  \E[q]{m(\rv \phi_1 \deformed M , F)} = \E[q]{m(\rv \Phi_1 , F)} = \trace \E[q]{\rv \Phi_1\rv \Phi_1^\transp} -2\bar\Phi^\transp _1F +\trace F F^\transp.
\end{equation}
Replacing \Cref{eq:disc-KL-data} in \Cref{eq:disc-KL-approach-momenta} leads to the gradient 
\begin{equation*}\label{eq:disc-KL-approach-update}
  \nabla_{ \mu_{t=0}}\mathcal E(q_{\mu_{t=0}}(\rv \phi_1)) = \tfrac 1 2 S^{-1}_{t=0} \mu_{t=0} - \left(  2\bar \Phi_1 -2 F \right).
\end{equation*}
Having this gradient, we minimize $\mathcal E$ w.r.t. $ \mu_{t=0}$ using a gradient descent algorithm.

%% file: experiments.tex
\begin{figure}
  \begin{center}
  \subfigure[Small Deformation\label{fig:synth-short}]{\includegraphics[height=.25\textwidth]{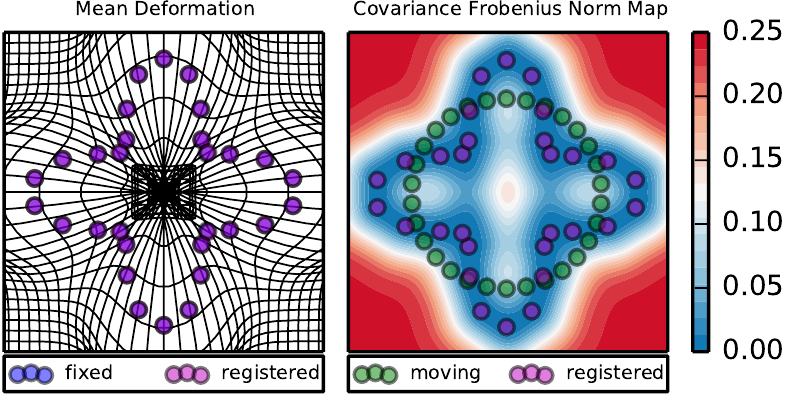}}
  \subfigure[Diffeomorphic with Geodesic Shooting\label{fig:synth-diff}]{\includegraphics[height=.25\textwidth]{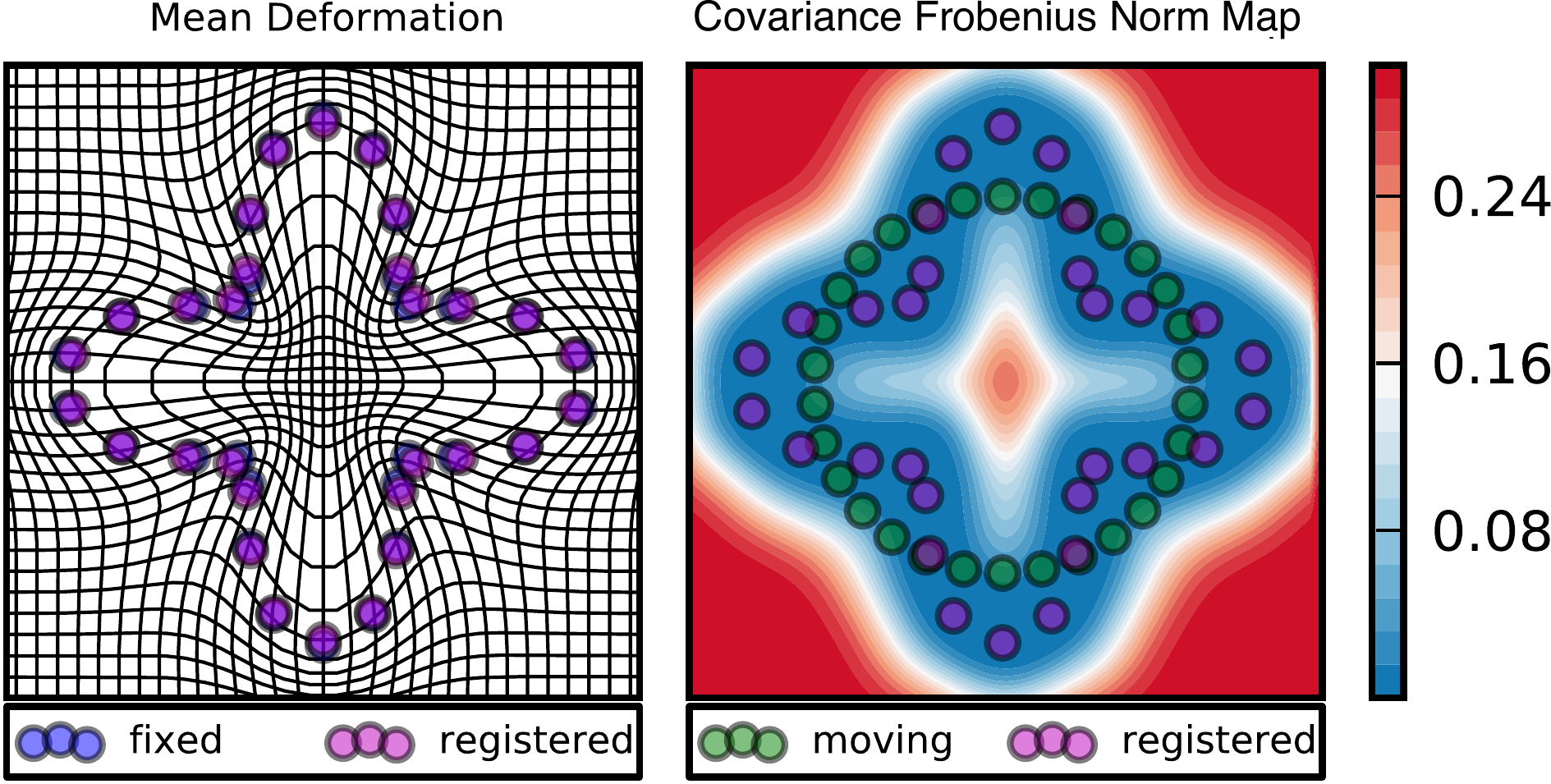}}
\end{center}
\caption{Comparison between small deformation and diffeomorphic registrations with equal parameter values, the \emph{uncertainty} is represented by the Frobenius norm of the covariance. The small deformation model has a smaller variance in general at the expense of a possibly invalid deformation field away from the landmarks.\label{fig:synth}\vspace{-15pt}}
\end{figure}

We are now in position to perform experiments using our probabilistic diffeomorphic registration algorithm. For all our experiments, we chose the covariance function
\begin{equation}\label{eq:cov-fun}
  \Sigma_{ts}(\vec x, \vec y)_{ij}=\delta({t-s})\left[\exp\left(-\tfrac{\|\vec x_i-\vec y_j\|^2_2}{2\sigma^2}\right)\right]_{ij}  \in \Re^{d\times d}
\end{equation}
where $\sigma^2$ is the model parameter. To conclude the specification of the model, we know with certainty that the starting point of the registration algorithm is the identity transform, hence $\phi_0(\vec x)\sim\GP(\vec x, \Sigma_{t=0,s=0}(\vec x, \vec y))$ and $\Sigma_{t=0,s=0}(\vec x, \vec y)=\vec 0$.

\subsection{Validity of the Locally Linear Approximation}\vspace{-8pt}
To test the validity of our GP model for diffeomorphic deformations we compared the GP through the LL method with one of the standard numerical solver for SDEs which does not assume a parametric density on $\phi$~\cite{Kloeden:1992vt}. We generated two sets of landmarks, as shown in \Cref{fig:synth}, a circle and one resembling a flower, both of radius $10mm$. Then we generated random initial velocity fields with the covariance function in \Cref{eq:cov-fun} with a range of $\sigma \in \{.1, 2, 5\}$. 
We sampled from the SDE in \Cref{eq:prop-sde} using the standard Euler-Maruyama method~\cite{Kloeden:1992vt} and then calculated the mean and covariance of the samples at the end time of the simulation. On the other side we calculated the mean and covariance at the same end time using the ODEs in \Cref{eq:phi-ll-mean,eq:phi-ll-2ndm}. After generating $100$ experiments per landmark set and $\sigma$ value, the mean arrival locations for both methods differed by $.5 \pm .02$ for $\sigma=.1$; $.1 \pm .003$ for $\sigma=2$ and $.012 \pm .0003$ for $\sigma=5$ all at least two orders of magnitude smaller than the radius of the datasets; the Frobenius norm of the difference between covariances was $11 \pm 1$ for $\sigma=.1$; $3 \pm .02$ for $\sigma=2$; and $.5 \pm .01$ for $\sigma=5$ which is small in comparison with the original variance of the points $74$. This shows good agreement between the LL and the Euler-Maruyama methods.

\subsection{Synthetic Registration Experiment}\vspace{-8pt}
In order to compare our diffeomorphic model with a stochastic short deformation model \cite{Simpson:2013ia}, we implemented our model and then registered the landmarks in the circle shown in green in \Cref{fig:synth} to those of the ``flower'' shown in blue. The results for the short deformation model are illustrated in \Cref{fig:synth-short} and those of the diffeomorphic in \Cref{fig:synth-diff}. It is noticeable that in the short deformation model the domain has been warped into a non-invertible deformation which is not possible in the diffeomorphic case~\cite{Ashburner:2011cp,Dupuis:1998ur}. 
We also show the uncertainty in the transform as modeled by the of the deformation field at each point. In \Cref{fig:synth-short,fig:synth-diff} it is noticeable how, as expected, the uncertainty is lower close to the landmarks and it grows as we move far away from them. Moreover, in both models the FC values are comparable, showing that the increased complexity of the diffeomorphic model has not increased the uncertainty in the results.

\subsection{Registration of Pre-operative and Intra-operative Images}\label{sec:real-data}\vspace{-8pt}
\begin{figure}
  \begin{center}
  \subfigure[MRI: Linear Registration]{\includegraphics[width=.2\textwidth]{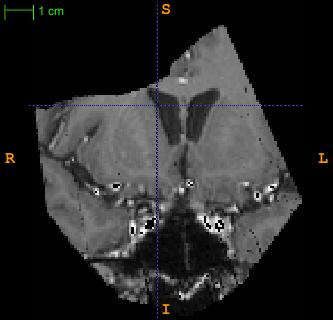}}
  \subfigure[Intraoperative Ultrasound]{\includegraphics[width=.2\textwidth]{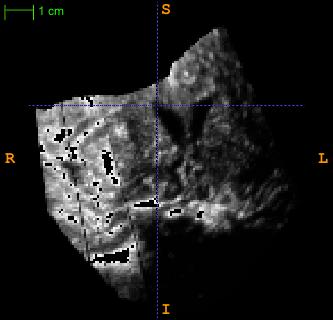}}
  \subfigure[MRI: Probabilistic Diffeomorphic Registration]{\includegraphics[width=.2\textwidth]{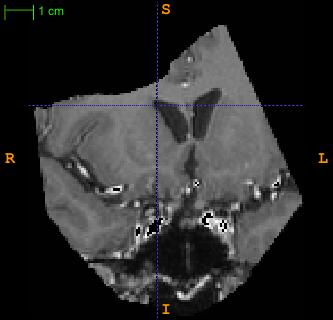}}
\end{center}
\caption{Registration of pre-operative and intra-operative images: (a) The pre-operative MRI linearly registered and projected onto the ultrasound space. (b) The intra-operative ultrasound image; and (c) the pre-operative MRI of (a)  registered to (b) using our algorithm were we show the warping according to the average registration field. The crosshair indicator shows how correspondence between a-b is not as accurate as the one using deformable registration (c).\label{fig:reg-us}\vspace{-15pt}}
\end{figure}
We illustrate the strength of our method in the case of multi-modal registration. We use publicly available images~\cite{Mercier:2012hd} which include $12$ clinical cases of brain tumor resection. For these cases T1-MRI images have been acquired pre-surgically, manually annotated with between 20 and 37 anatomical landmarks and a tumor delineation and then intra-operative 3D ultrasound (US) reconstructions were acquired for the same subjects before tumor resection. The same experts annotated the US images with the same landmarks as the MRI.  

\begin{figure}
  \begin{center}
  \subfigure[Validation \label{fig:loo}]{\includegraphics[height=.15\textheight]{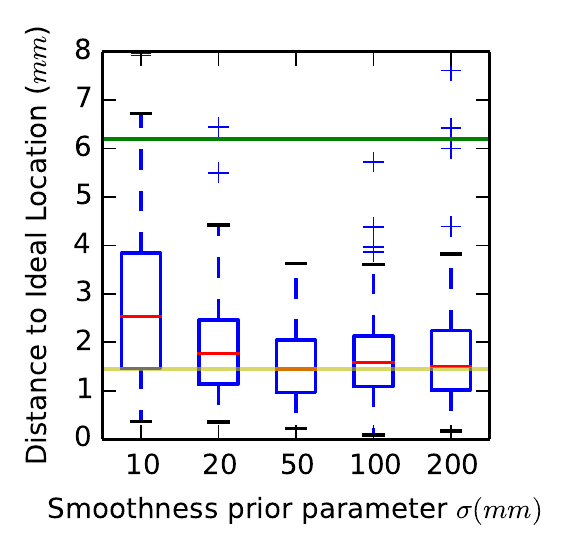}}
  \subfigure[Smoothing Comparison\label{fig:reg-us-un}]{\includegraphics[height=.15\textheight]{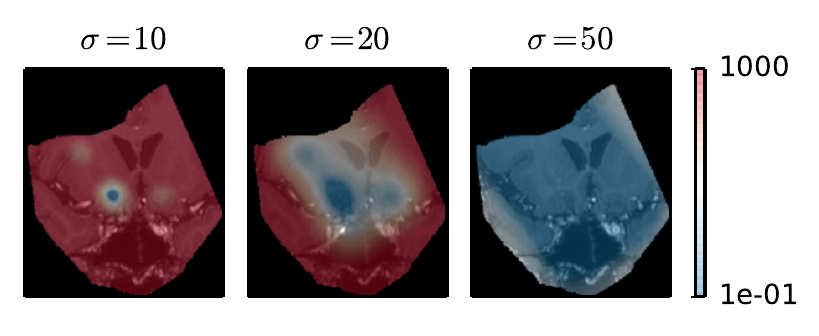}}
  \end{center}
  \caption{\textbf{(a)}: Evaluating registration accuracy against manually labeled landmarks using LOO (see \Cref{sec:real-data}).  The green line indicates the average pre-registration distance to the ideal location. The yellow line indicates the median distance to the ideal position of the best configuration, $\sigma=50mm$. \textbf{(b)}: The MRI of \Cref{fig:reg-us} warped according to the mean deformation of the probabilistic diffeomorphic registration using 5 different levels of smoothing. Overlapped on the warped image is the estimated uncertainty. As the smoothness of the prior increases the uncertainty of the warp diminishes spanning from a small neighborhood around the landmark to the rest of the image.\vspace{-12pt}}
\end{figure}


We tested the accuracy of our registration algorithm on areas were there is no explicit information. For this, we used a leave-one-out (LOO) validation. For each subject we took one of the landmarks out, registered all others and then measured the distance of the landmark that we left out with the solution that were obtained by including it in the registration.  We show the results in \Cref{fig:loo}. The results are over $12$ subjects with between $20$ and  $37$ landmarks per subject. We obtained the best results with $\sigma=50mm$. Priors with $\sigma<50mm$, were not able to move the left-out landmark to the ideal position and had increased variance. Priors, $\sigma > 50mm$, had a closer distance to the ideal location but an increased number of outliers.  Finally, we register these subjects using all the available landmarks and, through visual inspection, we are able the see that the deformable registration improves the image matching as shown in~\Cref{fig:reg-us}. Moreover, we also show how a prior enforcing a stronger smoothness constraint increases the certainty in of the registration in the whole image. We illustrate this in~\Cref{fig:reg-us-un} where the increase of the low uncertainty (blue) area of the image correlates with the increase of the smoothness parameter.

%% file: discussion.tex
In this paper we presented a probabilistic diffeomorphic registration methodology. By extending the usual diffeomorphic model of \citet{Dupuis:1998ur} from a deterministic ODE formulation to a stochastic one, we were able to include in our model the registration error, or uncertainty. To the best of our knowledge, this is the first algorithm proposing a probabilistic diffeomorphic registration using a parametric density of the diffeomorphic deformations including a numerical method to calculate the parameters.
Having presented our model, we devised an algorithm to implement it through a locally linear approximation to a parametric density. We successfully tested this approximation against usual methods for SDEs where a parametric density is not available. Then, we analyzed the performance of our algorithm in synthetic and human data. Our experiments showed that our algorithm produces good results. We measured this quantitatively through a LOO experiment as well as qualitatively by visual assessment of 12 registrations between MRI and US modalities.

%% file: wbir2014.bbl
\begin{thebibliography}{10}

\bibitem{Dupuis:1998ur}
Dupuis, P., Grenander, U.:
\newblock {Variational problems on flows of diffeomorphisms for image
  matching}.
\newblock Quarterly of Applied Mathematics (1998)

\bibitem{Joshi-Miller:2000}
Joshi, S.C., Miller, M.:
\newblock {Landmark matching via large deformation diffeomorphisms}.
\newblock TIP (2000)

\bibitem{Ashburner:2011cp}
Ashburner, J., Friston, K.J.:
\newblock {Diffeomorphic registration using geodesic shooting and Gauss--Newton
  optimisation}.
\newblock NImg (2011)

\bibitem{Beg-Miller-etal:2005}
Beg, M., Miller, M., Trouv{\'e}, A., Younes, L.:
\newblock {Computing large deformation metric mappings via geodesic flows of
  diffeomorphisms}.
\newblock IJCV (2005)

\bibitem{Simpson:2013ia}
Simpson, I.J.A., Woolrich, M.W., Cardoso, M.J., Cash, D.M., Modat, M.,
  Schnabel, J.A., Ourselin, S.:
\newblock {A Bayesian Approach for Spatially Adaptive Regularisation in
  Non-rigid Registration}.
\newblock In: MICCAI. (2013)

\bibitem{Markussen:2007hv}
Markussen, B.:
\newblock {Large deformation diffeomorphisms with application to optic flow}.
\newblock Computer Vision and Image Understanding (2007)

\bibitem{risholm2013bayesian}
Risholm, P., Janoos, F., Norton, I., Golby, A.J., Wells~III, W.M.:
\newblock Bayesian characterization of uncertainty in intra-subject non-rigid
  registration.
\newblock Medical image analysis (2013)

\bibitem{zhang2013bayesian}
Zhang, M., Singh, N., Fletcher, P.T.:
\newblock Bayesian estimation of regularization and atlas building in
  diffeomorphic image registration.
\newblock In: IPMI. (2013)  37--48

\bibitem{Kloeden:1992vt}
Kloeden, P.E., Platen, E.:
\newblock {Numerical Solution of Stochastic Differential Equations}.
\newblock Springer (1992)

\bibitem{Rasmussen-Williams:2006}
Rasmussen, C.E., Williams, C.K.I.:
\newblock {Gaussian Processes for Machine Learning}.
\newblock The MIT Press (2006)

\bibitem{Biscay:1996ho}
Biscay, R., Jimenez, J.C., Riera, J.J., Valdes, P.A.:
\newblock {Local linearization method for the numerical solution of stochastic
  differential equations}.
\newblock Ann Inst Stat Math (1996)

\bibitem{Mercier:2012hd}
Mercier, L., Del~Maestro, R.F., Petrecca, K., Araujo, D., Haegelen, C.,
  Collins, D.L.:
\newblock {Online database of clinical MR and ultrasound images of brain
  tumors}.
\newblock Med. Phys. (2012)

\end{thebibliography}
